# Bridge type classification: Supervised learning on a modified NBI dataset

Achyuthan Jootoo [1]; and David Lattanzi, M.ASCE[2]

1 PhD candidate, Department of Civil, Environmental, and Infrastructure Engineering, George Mason
  University, Fairfax, VA, USA, 22030; email: ajootoor@gmu.edu
2 Assistant Professor, Department of Civil, Environmental, and Infrastructure Engineering, George
  Mason University, Fairfax, VA, USA, 22030; PH (703) 993-3695; email: dlattanz@gmu.edu

## Authors' Note:

This work is published in Journal of Computing in Civil Engineering at the following link:
https://ascelibrary.org/doi/full/10.1061/(ASCE)CP.1943-5487.0000712

## Abstract

A key phase in the bridge design process is the selection of the structural system. Due to budget
and time constraints, engineers typically rely on engineering judgment and prior experience
when selecting a structural system, often considering a limited range of design alternatives. The
objective of this study was to explore the suitability of supervised machine learning as a
preliminary design aid that provides guidance to engineers with regards to the statistically
optimal bridge type to choose, ultimately improving the likelihood of optimized design, design
standardization, and reduced maintenance costs. In order to devise this supervised learning
system, data for over 600,000 bridges from the National Bridge Inventory database were
analyzed. Key attributes for determining the bridge structure type were identified through three
feature selection techniques. Potentially useful attributes like seismic intensity and historic data
on the cost of materials (steel and concrete) were then added from the US Geological Survey
(USGS) database and Engineering News Record. Decision tree, Bayes network and Support
Vector Machines were used for predicting the bridge design type. Due to state-to-state variations in material availability, material costs, and design codes, supervised learning models based on the complete data set did not yield favorable results. Supervised learning models were then trained and tested using 10-fold cross validation on data for each state. Inclusion of seismic data improved the model performance noticeably. The data was then resampled to reduce the bias of the models towards more common design types, and the supervised learning models thus constructed showed further improvements in performance. The average recall and precision for the state models was 88.6% and 88.0% using Decision Trees, 84.0% and 83.7% using Bayesian Networks, and 80.8% and 75.6% using SVM.

## 1. Introduction

The design and construction of a structure has several stages (Figure 1). In the first stage, the initial design requirements such as site constraints and usage demands are defined. In the next stage, preliminary design, the structural system and design materials are chosen (e.g.: steel trusses or concrete box beams). Because project budgets and time are limited at this phase, initial prototyping is generally reliant on the engineering judgment and prior experiences of the design engineer. This often means that engineers only consider a limited set of structural systems that seem most feasible, and explicit design optimization is not considered. The preliminary design forms the basis for a cost estimate and final design of the structure, and so choosing a sub-optimal structural system can lead to unnecessarily high project costs and other inefficiencies.

A computer-based decision support system could help guide preliminary design teams towards optimal structural designs by using initial design criteria as program input and rapidly providing initial projections of suitable structure types. These projections could also be used to avoid inefficient design alternatives, or to consider options that may not be apparent given the experiences and biases of a given engineering team. This paper presents a study on the capabilities of such a computational tool, one that uses supervised machine learning in conjunction with a newly compiled dataset for generating preliminary bridge structure prototypes.

## 1.1 Prior Work

Although there have been relatively few implementations of supervised learning in structural design and prototyping, analogous implementations have been successful in the medical field, as well as other fields within civil and structural engineering. In particular, the success of supervised learning for medical diagnosis, which is conventionally reliant on a doctor's professional experience, forms much of the basis for the work presented herein.

In medical diagnosis, a doctor must leverage both professional experiences and a patient's symptoms to provide an initial estimate of the patient's condition. This is, at its core, a nonlinear statistical problem that is well suited for machine learning. One of the key points noted by Bellazzi and Zupan (2008) is that the confidence of a physician in trusting the results of a machine learning-based disease prediction is higher if the model's decision making process is easily communicated to the doctor. This point is relevant for civil engineers who might trust a design prototype generated through machine learning prototype if it is more easily comprehensible. Detailed discussions about the various implementations of machine learning in medical diagnosis can be found in Harper (2005), Herskovits and Gerring (2003), Kononenko (2001), Kukar (2001), Lavrač (1999), Luciani et al. (2003), Potter (2007).

While there has not been much work in using supervised learning for structural design, machine learning has been previously used for other structural engineering applications, such as component level analysis (Hung and Jan 1999, Vanluchene and Sun 1990, Naeej et al. 2013, Sadowski and Hoła 2013, Chou et al. 2014). A survey of the machine learning and artificial intelligence applications to structural engineering over the years can be found in Adeli (2001) and Lu et al. (2012). There has been significant work on structural design optimization — a field closely related to machine learning — using evolutionary techniques  (Kicinger et al. 2005a, Kicinger et al. 2005b, Murawski et al. 2000). Machine learning, primarily in the form of pattern recognition, has also been widely used for statistical structural health monitoring (Worden and Dulieu-Barton 2004, Worden and Manson 2007). There are also a variety of studies beyond structural engineering. For instance, Abdulhai et al. (2003) and Chien et al. (2002) use machine learning for adaptive traffic signal design and bus prediction respectively.

## 1.2 Focus of this Research

When compared with the medical field and other civil engineering problem domains, there has not been a significant study into the use of supervised machine learning for structural design and prototyping. To address this need, presented herein is a study into the suitability of supervised machine learning techniques for generating preliminary bridge design predictions, using the National Bridge Inventory (NBI) database in conjunction with additional seismic and cost data for model training and validation. The primary contribution of this work stems from the analysis as to how to properly construct an accurate and efficient supervised classifier, or set of classifiers, capable of accommodating the breadth of design criteria across the United States. While the NBI dataset has been studied previously (Chase et al. 1999, Sun et al. 2004, Bolukbasi et al. 2004), no work so far investigates the use of NBI dataset for structural prototyping. To improve model performance, USGS seismic intensity data and historic material cost data from *Engineering News Record* (ENR 2016) were added to the NBI dataset, and the benefits of these datasets are considered as well. The next section describes the information modeling methodology used in this effort. Section 3 presents analyses and experiments conducted on the constructed classifiers. Section 4 presents overall conclusions and potential future research initiatives in this domain.

## 2. Methodology

Supervised machine learning (ML) encompasses algorithms that use data, usually in large amounts, to develop statistical models that can make predictions based on new instances of similar data. In this work, data from a variety of sources is first compiled into a new and more comprehensive dataset. Critically, the dataset used for model building is then cleaned of extraneous data using feature reduction techniques in order to prevent spurious correlations and improve computational efficiency. This reduced dataset is then used to construct and validate the selected classifiers. This overall approach is depicted in Figure 2.

### *2.1 Datasets Used for Analysis*

The implementation of any supervised ML algorithm requires a dataset from which to train and validate the resulting statistical models. Such a dataset is comprised of "instances", or data points, that are described through a set of numeric attributes. For bridge prototyping, an instance of a bridge design consists of the design criteria and attributes that governed the design. These parameters include, but are not limited to: geometric design requirements, details regarding the usage of the bridge, hazard and load parameters, and material cost data.

The NBI database is comprised of over 600,000 bridges with up to 110 attributes for each bridge, and contains most of the necessary information. Of the 110 attributes, a few key attributes are bridge length, number of spans, navigational vertical clearance, length of maximum span, deck structure type, sufficiency rating, approach road width, material type, design type, number of lanes on the bridge and average daily traffic. A list of all the attributes in the NBI database along with its explanation can be found in FHWA (2016).

Two key design considerations that are not a part of the NBI data set are design hazards, such as seismic loads, and material cost data. As a part of this research project, these attributes were added to the compiled NBI dataset. Seismic data for peak ground acceleration with a 2% probability of exceedance provided by USGS (USGS 2015) was used to represent the seismic hazard data for bridges in the continental 48 states. Seismic data for Alaska, Hawaii and Puerto Rico is currently outdated and hence was not included for analysis. The seismic data is provided at latitudinal and longitudinal coordinates at a resolution of 0.05° and a 2% probability of exceedance, and was associated with each bridge in the dataset at this resolution. Since seismic design based on USGS data has only been in practice since 1971, only the NBI bridge data for bridges constructed after 1971 have been considered for the models constructed in Section 3.4.2 and 3.4.4.

Data for the cost of concrete and steel for bridges were obtained from the *Engineering News Record* (ENR 2016) archives for the years of 1970 to 1991. The cost data included the cost of concrete and steel in 20 key cities across the United States (U.S.). For each bridge, the cost of the relevant material (e.g. steel for a steel bridge) for the nearest of the 20 cities is used. The cost used is for the year in which the bridge was completed since data could only be found for the year the bridge was built. The costs were converted to 2016 dollars before analysis. However, data for all the cities for each year is not available and hence, there is some missing data in the database. This 21-year span of data, selected due to the high variations in material costs during this period, was used to study the impact of material cost. Other associated environmental data considered in this study included annual data for rain, humidity and snow (Current Results 2016). The environmental data for rain, humidity and snow was available for each state separately and was not available for each bridge location in a state. Hence, this data was not added to the dataset, but was instead analyzed for linear correlations between the resulting model accuracies and the environmental parameters. The datasets developed in this research are available from Lattanzi (2016) for download and further processing.

*2.1.1 NBI Data Analysis*

The National Bridge Inventory classifies bridges into 23 types (FHWA 2016). A preliminary analysis of the NBI data, shown in Figure 3 shows the percentages and distribution of design types across the United States. Figure 3 shows that 41% of the bridges in the United

States are of type Stringer/Multi-Beam or Girder (as per NBI classification), 23% are culverts, 14% are slab bridges and 9% are Box Beam or Girders bridges. These bridge structure types account for over 85% of the bridges in the United States.

Further analysis was done for all bridges of the nation for which the material type was given to be steel. A pie chart for the percentages of different types of bridges is shown in Figure 4. The figure shows that for steel bridges, 77% of the bridges are of type Stringer/Multi-Beam or Girder, 11% are culverts, 7% use a truss system, 3% use a Girder and Floorbeam system and a small percentage of bridges are built using other design types. This pie chart indicates that knowing that the material type of a bridge can likely improve the chances of correctly predicting the design type since there are statistically fewer likely design types.

### 2.2 Dataset Reduction

Out of the 110 attributes for each bridge in the NBI dataset, not all attributes are useful for predicting the design type of the bridge as the purpose of the NBI dataset is to record the health of the bridge over its life cycle. Furthermore, many attributes from the NBI database would not be known at the design prototyping phase such as the sufficiency rating of the bridge. The complete dataset was therefore reduced prior to supervised learning testing and training.

From among all the attributes, a list of features that could affect the preliminary design of the bridge was first identified. Each selected feature had to satisfy one prerequisite: its value must be known, at least approximately, during the bridge prototyping stage. For this reason, parameters like sufficiency rating were not used for model development since their values cannot be estimated at the time of the structural design of the bridge. Using this prerequisite, and by eliminating unnecessary attributes (e.g. bridge address, latitude, longitude, county code, route number, etc.), the attribute list was reduced to 17 descriptors, an 85% reduction in dataset size.

After the first stage of feature reduction, the data was further reduced through attribute evaluation. This was performed by three different approaches. In the first approach, preliminary supervised learning models based on these 17 attributes were first constructed using a Bayesian Network model (Bouckaert 2004). Bayesian Network models were used since they are insensitive to noise and hence are well suited to the feature selection process (Witten et al. 2011). A single attribute was then removed from the dataset and a new model was constructed. This method is referred to as a "leave-one-attribute-out" strategy. If removal of an attribute did not affect Bayesian Network model accuracy by more than 1%, then it was considered as unimportant for prototype design prediction.

Feature selection techniques were then considered for use in data reduction. The comparative study of feature selection techniques in Ramaswami and Bhaskaran (2009) indicated that the information gain and chi-squared attribute evaluation methods performed well compared to other approaches, and were therefore used in this study. The information gain (Mitchell 1997) feature evaluation method measures the number of bits of information obtained for prediction of a class by knowing the presence or absence of an attribute in a data instance. The attributes with the highest values of information gains are considered the most relevant. Equations 1 and 2 provide a mathematical expression for information gain where Inf(S) denote information of the data S, $p_i$ is the proportion of S belonging to class I, values(A) depicts the set of values of attribute A and $S_v$ is the subset of S for which attribute A has value v.

$$Inf(S) = \sum_{i=1}^{c} -p_i \log_2 p_i \qquad (1)$$

$$InfGain(S, A) = Inf(S) - \sum_{v \in values(A)} \frac{|S_v|}{|S|} Inf(S_v) \qquad (2)$$

The chi-squared feature evaluation algorithm (Vryniotis 2016) ranks attributes by evaluating a metric which roughly measures the correlation between the class to be predicted and each attribute. This metric is evaluated as per Equation (3). In the equation, D represents the data which is being used for evaluating the metric, N is the observed frequency, E is the expected frequency, t is the attribute, c is the class, $e_t$ depicts whether the attribute value (term) t is in the data instance and $e_c$ depicts whether the data instance is of a particular class or not.

$$X^2(D, t, c) = \sum_{e_t \in \{0,1\}} \sum_{e_c \in \{0,1\}} \frac{\left(N_{e_t e_c} - E_{e_t e_c}\right)^2}{E_{e_t e_c}} \qquad (3)$$

In the chi-squared attribute evaluation, once the attribute scores were ranked, if the chi-squared score for any attribute was less than 70% of the preceding higher attribute score, that feature and all lower scored features were eliminated. After this, a leave-one-attribute-out analysis was performed on the attributes chosen by chi-squared evaluation algorithm to remove redundant attributes. Data reduction was performed independently by the leave-one-out-attribute out method and by a combination of chi-squared feature evaluation and leave-one-attribute-out (after feature evaluation) method. If any attribute was found to have impact as per leave-one-attribute-out method but was not chosen by the chi-squared evaluation, then that attribute was included. With regards to the information gain attribute evaluation, a clear cutoff for attribute selection could not be obtained. However, as seen in Table 1, the ranking of attributes was similar to the ranking obtained by chi-square, and served as validation for the overall feature selection process.

After all phases of data reduction, 4 NBI attributes were found to be statistically relevant to the bridge prototyping process: deck structure type, average span length, maximum span length and material type. Further, based on the leave-one-attribute out strategy, the seismic data attributes were also considered for the supervised learning models. The cost data, as obtained from the *Engineering News Record* (ENR 2016), was not found to be statistically useful for prediction. A discussion of the significance of cost data is presented in Section 3.4.3.

After the two stages of data reduction, the USGS 2% probability of exceedance seismic data was linked to each corresponding bridge by matching the latitude and longitude of the bridge to a corresponding acceleration in the USGS database, and then adding this acceleration as an attribute for each bridge instance. The cost data was linked to each bridge by making an assumption about the sourcing of materials for construction. The city center closest to each bridge was first found using the bridge's latitude and longitude. Next the cost of steel, the cost of concrete and the ratio of cost of steel and concrete from the closest city center were added as three separate attributes for the bridge.

Very few instances of missing data were observed in the NBI and USGS seismic intensity datasets. The NBI dataset however, had instances of improperly formatted data that resulted in errors while combining it with the USGS dataset. These instances were not included in the final dataset since there were several hundred such instances and each needed to be handled on a case-by-case basis. There was missing data in the cost dataset, and in such cases, the cost was assigned from the nearest source with an available material cost.

## 2.3 Supervised Learning Algorithms

There are a broad variety of supervised learning algorithms that can be implemented for developing statistical models. Each algorithm differs in the way it constructs a model and the way it is represented. An ideal supervised learning algorithm should not just be accurate in the way it classifies the instances; it should also be easy for a human to follow the decision making process i.e. comprehensible, capable of handling missing data and insensitive to noise in data. A human-comprehensible model is preferable since an engineer would be able to follow the logic behind the model, thereby increasing confidence in it, which is in line with the findings in the medical diagnosis community (Bellazzi and Zupan 2008). In this domain, the comprehensibility of the classifier model was instrumental in gaining acceptance of the algorithm from physicians. Of the available supervised learning algorithms, two were chosen for this study, that have been used successfully in medical diagnosis and are human comprehensible: Decision Tree and Bayesian Network classifiers (Herskovits and Gerring 2003, Luciani et al. 2003, Potter 2007). In order to evaluate the performance of these algorithms, the Support Vector Machine classifier was also used to construct models. The Support Vector Machine (SVM) (Tax and Duin 1999) classifier is considered to be one of the most consistent, robust, and accurate algorithms in modern machine learning and therefore serves as a benchmark for upper bound performance. However, SVM models are not easily interpreted by humans, making them less likely to gain acceptance in professional application scenarios.

*2.3.1 Decision Tree*

A decision tree is a step-by-step decision-making process in a tree structure. The decision tree is developed from a dataset by finding the attribute that gives the best probability of correctly predicting the class of the instance if only that attribute is used. Once this attribute is chosen, for each of its possible values, the attribute that maximizes the probability of classifying the instance is chosen. Then this process is repeated for all the attributes. There are several methods for choosing a best attribute, some of which are explained in Witten et al. (2011), Mitchell (1997). Figure 5 illustrates a mockup tree structure for selecting a bridge prototype.

Since the decision tree algorithm develops a model based on the data, it often develops a model which fits the training data very closely. As a result of this overfitting, the resultant model is susceptible to noise and can have poor performance on high variance test sets. To avoid this problem of overfitting, the decision tree is pruned (Witten et al. 2011). In this work, the J48 pruned decision tree (Weka 2015a) in Weka (Weka 2015b) was used. Weka provides two methods of pruning: reduced error and subtree raising (Witten et al. 2011). A confidence factor for pruning is also provided with lower values resulting in more pruning. Another parameter, minimum number of objects, controls the minimum number of instances per leaf. Larger values for the minimum number of objects reduces overfitting. In this paper, a sensitivity analysis (Section 3.3) was performed and, based on this analysis, decision tree models with a confidence factor of 0.35, minimum 2 objects per leaf, and 3 folds were used. Subtree raising led to better overall classifier performance compared to reduced error pruning and was thus used to construct all decision tree models.

*2.3.2 Bayesian Network Classifier*

A Bayesian network is a directed acyclic graph where each node represents a stochastic variable and arrows/arcs represent probabilistic dependency between a node (where the arrow points) and its parents (where the arrow begins). The probability of each possible class is determined from the learned network using the chain rule and the class with the maximum probability is assigned as the class for the instance (Bouckaert 2004). Details about the implementation of these algorithms can be found in Cheng and Greiner (1999), Mitchell (1997), Witten et al. (2011). Figure 6 represents a mockup network structure for selecting a bridge prototype.

The algorithm searches the space of graphs in order to develop a Bayes network based on the data and then computes the conditional probability values for each dependency. In this work, the Bayesian Network algorithm (Bouckaert 2004) was used. The available parameters for searching networks and computing probability were first tested. Based on this sensitivity study (Section 3.3), Bayesian Networks with the K2 algorithm (Cooper and Herskovits 1992) for searching network structures and the Simple Estimator (Bouckaert 2004) for developing the conditional probability tables were used.

*2.3.3 Support Vector Machine*

Conceptually, an SVM constructs the best hyperplanes through a dataset in order to divide the data into different classes. For a set of points, an SVM finds a hyperplane (Witten et al. 2011) which is at a maximum distance, M, from the closest data points of each class it divides. The SVM performs an optimization to maximize this distance M. The points which are at a distance M and closest to this hyperplane are called the support vectors. In this work, SVM (LibSVM implementation in Weka) with a radial basis function was used to develop supervised learning models. Other kernel functions such as polynomial and a sigmoid kernel were tested but yielded poor results and hence were not used.

It is important to note that while these three algorithms (DT, BN, SVM) were compared throughout the study, the aim was not to determine the best classifier, as the field of supervised machine learning is advancing rapidly and it is difficult to definitively determine the most accurate approach at any given time. Instead, the intent was to explore the possibility of using supervised machine learning in any capacity for the bridge prototyping process, and to explore the critical aspects of doing so.

*2.3.4 Classifier evaluative metrics*

Classifier performances were evaluated in several ways. Overall prediction accuracy for structure design type (or recall) along with the precision was the most important metric, as was the variance in prediction accuracy and precision from state to state. All models were tested and validated using 10-fold cross validation (Witten et al. 2011), with 90% of each dataset used to train the model, followed by testing on the remaining 10% of the data. This process is then repeated 10 times (folds) with a different 10% used for testing each time. The exception is the one state hold-out analysis of Section 3.2. Lastly, the models constructed on the complete national database were compared against a One Rule (OneR or 1R) classifier (Witten et al. 2011) that selected bridge prototypes based on only one attribute, in order to assess the need for sophisticated supervised learning.

# 3. Experiments and Results

The initial phase of the experiments was the reduction of the dataset (Section 3.1). After the data was reduced to only the critical attributes, a series of supervised learning studies was performed. The first study looked at building a supervised learning model using the entire national database, in order to determine if it was possible to develop a simple unified model for predicting the design type of all the bridges in the U.S. (Section 3.2). A brief sensitivity analysis discussing the impact of various parameters on the models is provided in Section 3.3. Supervised learning models given each individual state's bridge inventory were then constructed using Weka and analyzed, in order to develop more fine grained models that captured the design preferences of each state (Section 3.4). For the individual state studies, the impact of including seismic attributes (Section 3.4.2) and material cost attributes (Section 3.4.3) was investigated. Each individual state dataset was also resampled to study the impact of this technique on model performance (Section 3.4.4).

The last experiment studied model correlations with external factors that may have impacted classification accuracy, but would not necessarily be explicitly considered at preliminary design. For instance, states with a relatively high average humidity are less likely to consider design alternatives that are vulnerable in such conditions, but this is difficult to model numerically for supervised learning and hence, a correlation study was performed. The parameters studied were humidity, average daily temperature, average rainfall, and average snowfall (Section 3.5).

## *3.1 NBI Dataset Cleaning and Reduction*

After the first step of initial feature reduction, 17 relevant attributes remained out of an initial set of 110. A new feature, called average span length was developed and added to the data of each bridge. Next, the second round of feature selection was carried out by both the approaches outlined in Section 2.2. The first approach, the leave-one-attribute-out analysis, reduces the number of features from 17 to 4. Removal of any one of these remaining four attributes results in significant drop of over 1.5% for the Bayesian Network model. These four attributes were maximum span length, average span length, deck structure type and material type. Chi-squared feature evaluation resulted in different key attributes for different states, but after performing a leave-one-out analysis, the four attributes determined via the initial leave-one-attribute-out process were always the best attributes for determining the classification accuracy. In Table 1, the chi-squared score of each attribute is shown in brackets. It was observed that there was almost always a large decrease in chi-squared evaluation score after the first attribute. Hence, the 70% elimination criterion was only used for the second highest scoring attribute and beyond.

Note that for the state of Kansas, Deck structure type, which scores highly for the other states, is less than the previous attribute score by about 34%. However, upon adding this attribute to the supervised learning models for Kansas, the model performance improves significantly and hence this was considered for the supervised learning models of Kansas.

For the seismic intensity attribute and the cost data attributes, the chi-squared evaluation scores were very low. However, the leave-one-attribute-out method results indicated that including seismic data improved the predictive accuracy of the models significantly (Section 3.4.2). Hence, the seismic data was considered as another attribute for the models. In contrast, the cost data did not have any impact on the model predictive accuracy. However, to thoroughly examine the impact of the cost data, a more detailed discussion, as outlined in Section 3.4.3, was performed.

### 3.2 Preliminary Database Analysis

The first model was developed using data from all of the bridges across the United States concurrently. The model was first built using the One Rule (OneR) classifier, which resulted in an overall classification accuracy (recall) of 61.5% and a precision of 57.1%, representing a lower bound for statistical analysis. Using a Bayesian Network classifier, a recall of 74.9% with a precision of 73.5% was achieved. A J48 Decision tree further improved the recall to 82.5% and the precision to 81.1%. Both of these models performed significantly better than the One Rule classifier, illustrating the benefits of supervised learning in this domain. However, to evaluate the variance in accuracy from state to state, a second set of models was developed.

In this set, classifiers were trained using the nationwide NBI database. For each model, 49 states were used for model training and one was held out for model testing. Both BN and DT models were constructed, and the results are shown in Table 2. Analyses were also conducted by partitioning the whole dataset based on the material type and the deck structure type. In these analyses, separate models were developed for all bridges of a particular material type (or deck structure type). The results obtained were characterized by poor precision and recall, and high variance, and have not been tabulated in this paper.

The high performance variance between states (overall standard deviation of 13.7 for precision and 15.2 for recall in DT, and 17.0 for precision and 17.9 for recall in BN) indicated that a single nationwide model is not sufficiently reliable for use as a recommendation tool for engineers. This is likely due to the variations in design codes for each state that would induce inherent changes in the design choices. In addition to this, the results suggest that a single nationwide classifier is not able to capture regional differences in environmental considerations. As a result of this initial study, a set of per-state models were constructed and explored.

### 3.3 Sensitivity Analysis of Supervised Learning Algorithms

In order to obtain the best set of parameters for the supervised learning algorithms used, a sensitivity analysis was performed. For Decision Trees, different values for three parameters were tried out. Different values of confidence factor (0.15, 0.25, 0.35 and 0.45) were tested and the impact of using reduced error pruning was evaluated. The minimum number of objects per leaf was set to 2, 4 and 6. Although the performance variations for the parameters tested were within 3%, the best performance for a DT was obtained for a confidence factor of 0.35, minimum 2 objects per leaf and no reduced error pruning and these values were used for the DT classifier in this research.

In the case of Bayesian Network, different algorithms for searching the network structure were tested. The LAGD Hill Climber (Abramovici et al. 2008), Repeated Hill Climber , Tabu Search (Glover 1989) and K2 search (Cooper and Herskovits 1992) were used and the best performance was obtained for the K2 search. Different estimators for computing the conditional probability tables were tested and the best result was obtained for the Simple Estimator. The performance variations for the different parameters in Bayesian Network were within 2%.

For the SVM, different kernel functions were experimented with. The use of different kernel functions showed performance variations of over 10% which indicates that the choice of the correct kernel function is critical to the performance of an SVM classifier. The radial basis function kernel, sigmoid kernel and polynomial kernel were used with the best performance obtained for the radial basis kernel.

### 3.4 Individual State Analysis

#### 3.4.1 Models using attributes from NBI dataset only

Table 3 tabulates the bridge prototype precision and recall from the models learnt and tested on the data of each state. From the table, it is apparent that the model performance is significantly different for each of the classifiers. The reason is that the three algorithms use the same data to choose potential models in different manners and thus the models that these algorithms learn differ from each other.

From Table 3 it can be seen that, across classifiers, model performance was consistently highest for the states of Georgia, Minnesota, and Mississippi. From the table it can also be seen that states with a higher risks of earthquakes, such as California and Washington, resulted in models with relatively poor performance, particularly for the Bayesian Networks. However, this is not the only reason for poor accuracy and seismically inactive states like Pennsylvania, Idaho and New Jersey also perform relatively poorly.

Overall, the recall for all the different classifiers was on an average higher than their precision. In case of Bayesian Network, the recall is higher by 1.7% whereas in case of Decision Trees, the precision is lower by 2.5%. For SVM, the difference was significantly higher at 5.6%.

It can be noticed that prior knowledge of material type in the models in Table 3 resulted in higher classification accuracy than results without known material type (Table 4). When material types are not known, the highest accuracies are from the decision tree for Mississippi and the Bayesian network for Georgia. Once more it can be noted that the classification accuracy is poor for states with high probability of earthquakes.

Table 5 shows the change in accuracies of the models of each state when they are developed with and without the use of the material type attribute. The average decrease in recall was 7.3% for DT, 7.6% for BN and 6.2% for SVM. The reduction in precision for the models was higher than the drop in recall. The decrease in precision was 8.6% for DT, 8.5% for BN and 7.7% for SVM. The highest reductions were observed for Pennsylvania and Vermont with over 15% decrease in recall and precision for all models in both the states. High decreases in accuracies were also noticed in the states of Illinois, Georgia and Ohio.

The average recall and precision across state models (4 attributes) for the DT classifier was 83.5% and 81.0%, with a standard deviation of 7.7% and 8.7% respectively. The average recall and precision of the BN model was 82.9% and 81.2%, with a standard deviation of 8.4% and 9.6% respectively. For the SVM model, recall and precision were 80.0% and 74.4%, with standard deviations of 9.0% and 11.7% respectively. In general, both the BN and DT models performed as well, if not better, than the benchmark SVM classifiers, though there were variations from state to state. It is also worth noting that the while the average recall is higher than precision, its standard deviation is also slightly lower than that of precision. Overall, this suggests that engineers can select a classification algorithm based on alternative criteria such as model comprehensibility and visualization, as recommended in medical diagnosis literature (Bellazzi and Zupan 2008).

In both the cases with material type data and without material type data, we note that the both recall and precision are noticeably low in states with high seismic activity. The reason for this was not explicit in the results, but a possible explanation could be that in these states, different locations within the state would have different seismic intensity. As a result, bridges with similar attributes could have different designs. This results in conflicting data in which identical (or very similar) parameters can result in one design type for one bridge and another for some other bridge resulting in poor performance for states with high seismic activity. Overcoming this deficiency was the focus of the following study.

*3.4.2 Models developed after incorporating seismic intensity from USGS data*

Experiments performed after the USGS 2% probability of exceedance seismic data was incorporated as a new attribute yield the results shown in Table 6. Overall model prediction accuracy (recall) increased by 2.0%, 0.8% and 0.8% for DT, BN and SVM respectively. In contrast, the precision increased by 2.5% for DT, 1.6% for BN and 1.1% for SVM. For most states the effect of adding this parameter lead to increases in accuracy of the model but there are instances of a few states showing a slight decrease in accuracy. The state of Ohio shows a slight decrease in recall for SVM models and a significant decrease in precision. On the other hand, noticeable improvements in model performance were observed for the seismically active states of Oregon and Washington as shown in Table 6. Although the recall for Oregon decreases slightly, the precision increases significantly. South Carolina, which also has seismically active areas, shows a 9.9% improvement in both precision and recall for the DT model but lesser improvements for BN and SVM. The performance of the classifiers for California did not improve significantly due to large portions of California being designed for seismic loads. Hence, the additional seismic activity data does not aid the classifier significantly. From these results it can be seen that the USGS data showed noticeably higher improvements in prediction accuracy in the states with relatively higher variations in seismic intensity in addition to slight improvements across all states.

Table 7 shows the difference in the number of correct predictions (using decision tree) for the most common design types in South Carolina. It can be seen that the initial model without seismic attribute predicted only 73.1% of the bridges correctly for "Stringer/Multi-beam or Girder" bridges and predicted 87.8% of the "Slab" bridges correctly. Once the seismic dataset was included, the correct predictions for these bridge types increased to 94.6% and 92.7% respectively. This increase in correct predictions improves the precision and recall of the model for South Carolina by 9.9% each. For the other design types, the differences in number of correct prediction were not significant.

*3.4.3 Models developed after incorporating historic material cost data from ENR*

The aim of using cost data was to use it as a metric for estimating the cost of the bridge if it were constructed with each material type (steel, concrete and timber). The approximate cost of a bridge should, in theory, heavily influence prototyping choices, and therefore model prediction accuracy.

The dataset was expanded to integrate cost data from bridges constructed between 1970 and 1991, along with both the NBI-derived data set and the seismic accelerations. In order to test the engineer's bias of cost data being useful, a study was performed. There were two goals in this study. The first was to determine the overall potential significance of the cost data in the model building process. The second was to explore if cost data could serve as a suitable replacement for the material type in the supervised learning models. Initial experiments using the chi-squared metric were performed to check the statistical importance of the cost attributes. Cost data was incorporated by including the steel and concrete costs as separate attributes. These experiments ranked the cost attributes as among the least important attributes in the dataset.

Next, a leave-one-attribute-out study was performed to evaluate the usefulness of the cost attribute. In this evaluation, models were built with several combinations of attributes. All models used the following four attributes: seismic data, maximum span length, average span length and deck structure type. One set of models were developed with the material type and cost data as additional attributes, in order to assess if the cost data provided additional information beyond the material type. A second set was developed using the cost data as a stand-in replacement for the material type. A third set was developed with material type data and no cost data for comparative purposes. There was little to no change in precision and recall between models constructed using the first and third data sets. The recall for the second set of models was noticeably lower than for the third set, indicating that the cost data as compiled was not a suitable replacement for material type. As such, the results of this study are not tabulated here. However, it can be inferred that the cost data does not add much to the knowledge that is already contained in the dataset. There are multiple possible reasons for this:

a) The assumption that the material for each bridge is sourced from the closest city center might not necessarily be correct.

b) The cost data incorporated into this dataset is the cost of the material in January of each year from 1970 to 1991. The cost that should actually be used is the cost that was used during the construction of the bridge. However, the dataset uses cost at the point when the bridge was constructed instead of an average concrete cost across the time span during which the bridge was constructed.

c) The purpose of incorporating the cost of steel and material was that the cost would be an approximate indicative index of the overall cost of the bridge. However, the actual relation between the cost of the material and the overall cost of the bridge is more complex than is represented through these models.

### 3.4.4 Models developed using resampling filter

In the NBI dataset, the bridges can be classified into 21 design types. However, each of these types is not equally prevalent, which can potentially bias a supervised learning algorithm towards the more common design types. To counter this effect, resampling was performed to produce a random subsample of the dataset (Weka 2015c). The subsample generated is biased in order to produce more data points for the less common design types by sampling with replacement. It was expected that as a result of this resampling bias, which offsets the supervised learning bias, the results would improve. A study was performed to examine the effect of the resampling bias on the performance of models that combined 4 NBI attributes with seismic intensity data. Experiments were performed with biases of 5%, 10%, 20% and 35%. The biggest improvement in model performance was obtained with a bias of 10%. The results of experiments performed after resampling the data (with 10% bias) are provided in Table 8. While noticeable improvements were obtained in the recall for Oregon and Washington, the increase in precision was even more marked (see Table 8). Some states also showed decreases in accuracy after resampling. This was noticeable in states with very good model performances such as Georgia, Minnesota and Mississippi. Overall, the results indicated that state models with a range of seismic design accelerations benefited from resampling, but low seismic activity state models did not significantly benefit. The average increase in model accuracy for DT models (5.1% recall and 7.1% precision) and BN models (1.1% recall and 2.5% precision) was noticeable, although benchmark SVM models saw slightly poorer performance (-1.9% recall and -1.6% precision). In general, the decision to resample data should be done on a case by case basis, based on data variance within a given state.

### 3.5 External parameters affecting classification accuracy

In addition to the features studied for the supervised learning models, four external parameters were analyzed for correlation with algorithm classification accuracy: average annual temperature, average annual humidity, average annual rain and average annual snowfall. Structures built in high humidity and precipitation regions are more prone to corrode which could alter the design trade-offs between steel and concrete. Further, high humidity and precipitation areas may tend to have fewer wood structures because humidity can lead to decay of wood. Extreme temperature variations lead to higher thermal strains in structures thus bridges built in such regions would have to account for such variable strains. If the data for rain, snow, temperature and humidity is available for the location of each bridge, then this data could be integrated with the NBI dataset and an analysis could be performed. However, the data for these four external parameters was only available for a few cities in each state and not for each bridge in the NBI database. Therefore, this data could not be integrated with the NBI datsaset and was not used as part of the analysis so far. Hence, to evaluate the relation between these parameters and the classification accuracy, a correlation study was performed.

The correlation between these parameters and the classification accuracy of state level models developed via resampling (Section 3.4.4) was evaluated. It was seen that classifier accuracy had the maximum linear correlation with temperature. However, even this correlation was fairly weak and not significant. The correlation coefficient for most of the other parameters and their combinations was observed to be close to zero. This indicated that the environmental conditions, as represented in this study, were not statistically relevant from a design perspective.

### 3.6 Results

The best model prediction accuracy of 97.1% was obtained for the state of Mississippi using DT and the average recall and precision of all the states (on resampled data including seismic intensity as an attribute) using DT was 88.6% (standard deviation: 5.1%) and 88.0% (standard deviation: 7.1%) respectively. Average recall and precision for BN (resampled data) was 84.0% and 83.7% with standard deviations of 7.6% and 8.2% respectively. SVM had a recall of 80.8% with a standard deviation of 7.9% and precision of 75.6% with a standard deviation of 10.6% (for data including seismic activity but without resampling). It was also noticed that in addition to the increase in recall for the DT and BN models, resampling also showed marked improvements in precision. Seismically active states have a broad range of potential seismic accelerations resulting in a wider variety of design choices for similar design conditions. Hence, the seismically active states of California, Oregon and Washington performed poorly, but their performance improved after including seismic intensity as attribute and resampling. These results show that while the models developed are not perfect, their performance in bridge prototype prediction is reasonably good.

From the model performance discussed in Section 3.4, it was noticed that the performance of the Bayesian Network models was less affected by making slight changes to the data. One of the key reasons for this is that the Bayesian Network models are more capable of handling noisy data. This also means that if a few bridges were designed by using a wrong design type (eg. the best design type was to use a slab bridge but the designer used a Tee beam design) will not change the model accuracy. Hence, this algorithm is more robust than the Decision Tree algorithm which is quite sensitive to noise in data. While this study does not intend to compare the two algorithms, it is worth noting that in cases where robustness to noisy data is required, Bayesian Networks would work better than Decision Trees.

## 4. Conclusions and Future Work

This research implemented supervised learning methods on the National Bridge Inventory dataset in an effort to classify structural design types based on preliminary design parameters. The dataset was modified by adding data about seismic activity and cost of concrete and steel. Attribute evaluation methods were used for selecting the best attributes for the learning algorithms. The supervised machine learning methods were used on the final dataset to predict the design type of the bridge. The effects of resampling the data on the model accuracies were investigated. The impact of including seismic activity data and cost data was also examined. Different supervised learning algorithms were used in this study. Since SVM is considered as a very good classification algorithm in the domain of supervised learning, it was used as a benchmark to evaluate the performance of the other two algorithms used in this study (Decision Trees and Bayesian Networks). The results showed that the performance of DT and BN was as good as, if not better than, SVM across all states for all the experiments conducted.

The experiments performed indicated that a single model capable of predicting the bridge prototype of all the bridges in the U.S. had excessive variance in the prediction accuracy, and therefore could not be considered reliable for design prototype recommendations. The models developed using individual state data performed much better, with improved prediction accuracies. The classification models developed for the states showed significant variance in their performance depending on the attributes included and from state to state. The model precision and recall improved significantly when material type was included as an attribute and also showed slight improvements upon inclusion of seismic intensity as an attribute. It should be noted that information about the material type is not always known at the preliminary design stage. Inclusion of the historic cost data did not improve the results significantly since the relation between the cost of materials and the cost of the bridge is extremely complex and the algorithm cannot model this relationship without case-by-case data on the cost of the bridges. Thus the cost of the materials, as represented in this study, provided no additional information to the algorithm in the process of deciding the best bridge type.

Future studies will explore better and more detailed representations of construction and material costs for the learning models. This work will also be extended to leverage the predictions from these supervised learning models to develop bridge final designs as per the design codes through advanced computational techniques such as topology optimization and optimization techniques such as evolutionary computation. The different predictions made with different material types will be used as seed points for the evolutionary algorithm.

# List of Figures

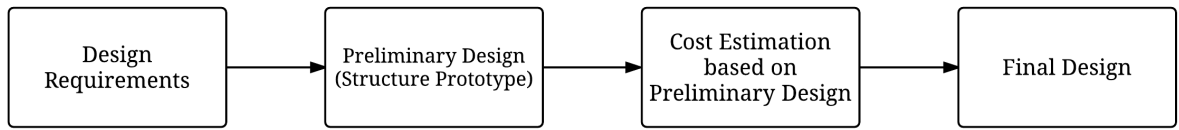

**Figure 1. Stages in the design and construction of a structure**

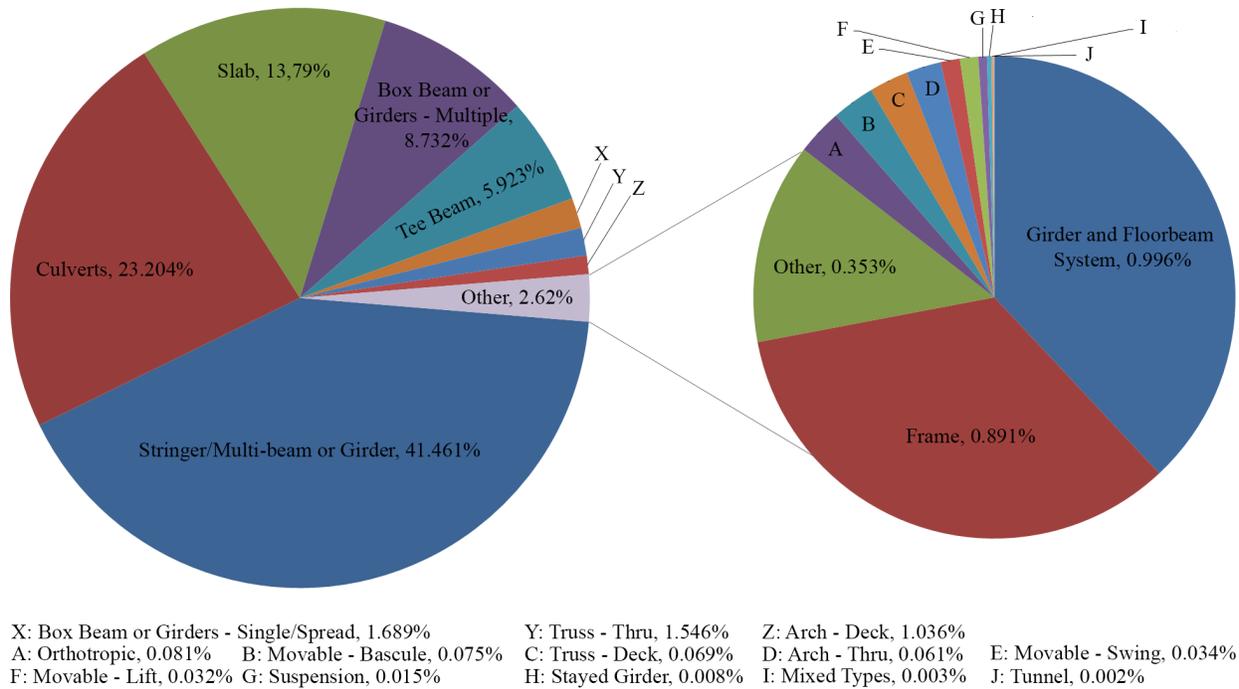

X: Box Beam or Girders - Single/Spread, 1.689%    Y: Truss - Thru, 1.546%    Z: Arch - Deck, 1.036%
A: Orthotropic, 0.081%    B: Movable - Bascule, 0.075%    C: Truss - Deck, 0.069%    D: Arch - Thru, 0.061%    E: Movable - Swing, 0.034%
F: Movable - Lift, 0.032% G: Suspension, 0.015%    H: Stayed Girder, 0.008%    I: Mixed Types, 0.003%    J: Tunnel, 0.002%

**Figure 2. Breakdown of bridge design types in the U.S.**

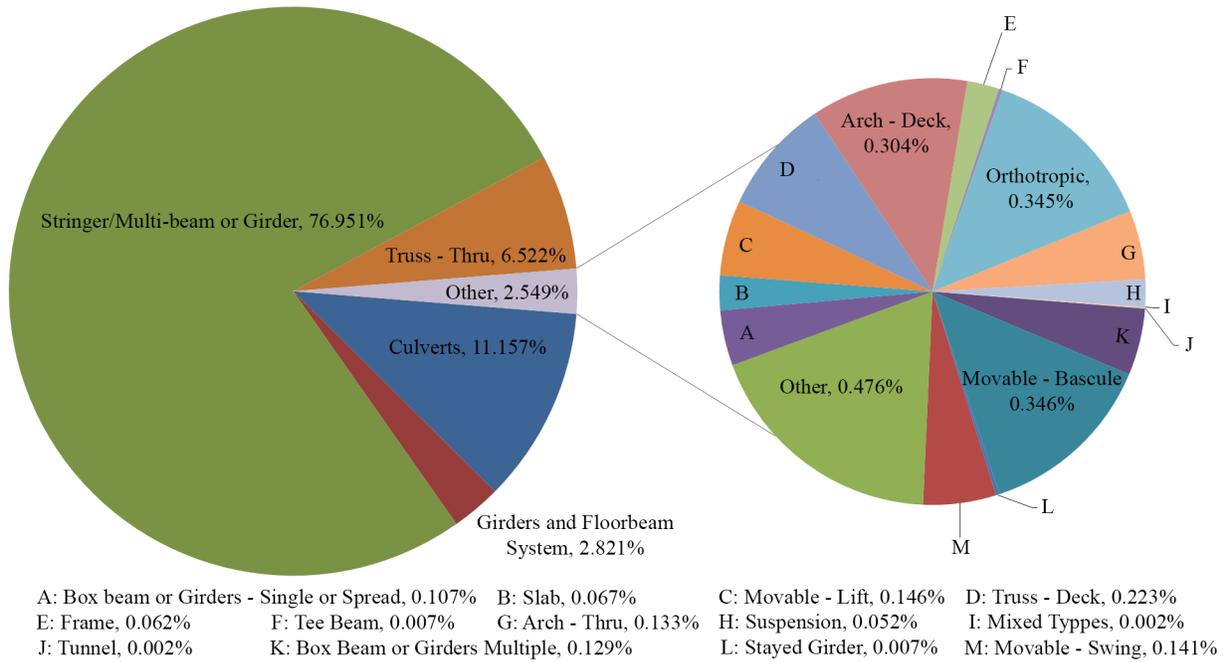

A: Box beam or Girders - Single or Spread, 0.107%  B: Slab, 0.067%  C: Movable - Lift, 0.146%  D: Truss - Deck, 0.223%
E: Frame, 0.062%  F: Tee Beam, 0.007%  G: Arch - Thru, 0.133%  H: Suspension, 0.052%  I: Mixed Typpes, 0.002%
J: Tunnel, 0.002%  K: Box Beam or Girders Multiple, 0.129%  L: Stayed Girder, 0.007%  M: Movable - Swing, 0.141%

**Figure 3. Breakdown of steel bridge design types in the U.S.**

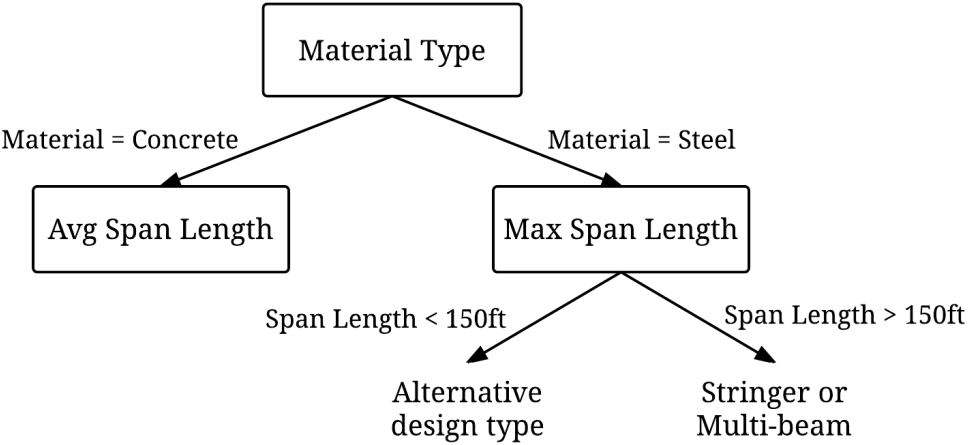

**Figure 4. Example Decision Tree for bridge design prototyping**

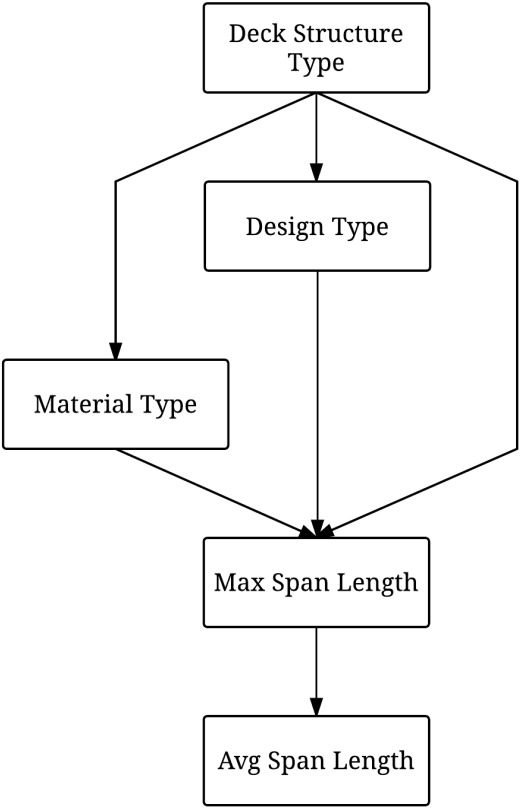

**Figure 5. Example Bayesian Network for bridge prototyping**

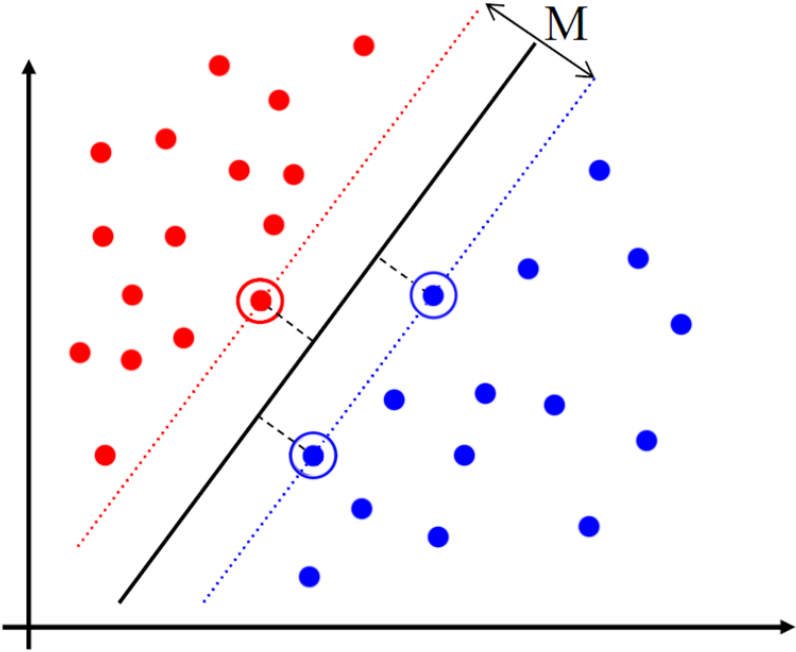

**Figure 6. Illustration depicting the working of an SVM**

# List of Tables

**Table 1 List of Best 6 Attributes Based on Chi-Squared and Information Gain feature selection. Asterisks (\*) indicate attributes selected for supervised learning models**

| Texas | | Kansas | | Georgia | | California | |
|---|---|---|---|---|---|---|---|
| Chi-square | InfoGain | Chi-square | InfoGain | Chi-square | InfoGain | Chi-square | InfoGain |
| Average Span Length* (482672) | Average Span Length* (1.30) | Average Span Length* (97055) | Average Span Length* (1.19) | Average Span Length* (113311) | Average Span Length* (1.52) | Max. Span Length* (57646) | Max. Span Length* (1.03) |
| Max. Span Length* (89006) | Max. Span Length* (1.14) | Material Type* (49317) | Material Type* (1.18) | Material Type* (36123) | Max. Span Length* (1.31) | Avg. Span Length* (36746) | Avg. Span Length* (0.89) |
| Material Type* (66641) | Material Type* (0.73) | Max. Span Length* (41666) | Max. Span Length* (0.92) | Deck Structure Type* (35596) | Deck Structure Type* (1.25) | Deck Structure Type* (31838) | Material Type* (0.77) |
| Total Length (50461) | Total Length (0.66) | Total Length (25597) | Total Length (0.61) | Max. Span Length* (35715) | Material Type* (0.88) | Material Type* (31659) | Total Length (0.50) |
| Deck Structure Type* (35449) | Deck Structure Type* (0.38) | Deck Structure Type* (16902) | Deck Structure Type* (0.46) | Total Length (16412) | Total Length (0.84) | Navigational Vertical Clearance (20313) | Deck Structure Type* (0.46) |
| Number of Spans (11251) | Number of Spans (0.16) | Number of Spans (8774) | Number of Spans (0.24) | Navigational Vertical Clearance (12503) | Min. Vertical Clearance (0.23) | Total Length (17585) | Min. Vertical Clearance (0.37) |

**Table 2 Precision and Recall of Decision Tree and Bayesian Network models on nationwide models of different states**

| Attributes Used | Material type, Maximum span length, Average span length, Deck structure type | | | |
|---|---|---|---|---|
| **State** | **Decision Tree (%)** | | **Bayesian Network (%)** | |
| | **Precision** | **Recall** | **Precision** | **Recall** |
| California | 62.0 | 51.1 | 43.3 | 44.7 |
| Florida | 79.7 | 75.4 | 79.4 | 78.0 |
| Georgia | 88.5 | 85.3 | 88.6 | 81.8 |
| Iowa | 82.6 | 82.5 | 80.1 | 82.1 |
| Maine | 75.1 | 76.0 | 76.1 | 74.8 |
| Minnesota | 87.2 | 85.4 | 84.0 | 85.1 |
| Mississippi | 88.3 | 76.8 | 90.1 | 84.0 |
| Nevada | 70.1 | 69.8 | 68.6 | 65.0 |
| Oregon | 55.0 | 39.7 | 48.0 | 34.9 |
| Pennsylvania | 49.0 | 53.3 | 47.4 | 52.7 |
| Virginia | 77.3 | 78.9 | 81.4 | 82.5 |
| Washington | 54.5 | 49.7 | 53.1 | 46.0 |

**Table 3  Accuracy of Decision Tree (DT), Bayesian Network (BN) and SVM models on different states (4 attributes)**

| Attributes Used | Material type, Maximum span length, Average span length, Deck structure type | | | | | |
|---|---|---|---|---|---|---|
| **State** | **BN (%)** | | **DT (%)** | | **SVM (%)** | |
| | **Precision** | **Recall** | **Precision** | **Recall** | **Precision** | **Recall** |
| California | 73.9 | 75.9 | 77.7 | 79.4 | 76.3 | 78.3 |
| Florida | 81.0 | 86.0 | 82.0 | 83.7 | 75.2 | 81.8 |
| Georgia | 96.2 | 96.9 | 95.5 | 96.2 | 90.5 | 93.3 |
| Illinois | 90.4 | 92.1 | 88.7 | 91.0 | 85.2 | 89.4 |
| Maine | 79.2 | 83.1 | 80.4 | 84.2 | 76.2 | 83.2 |
| Minnesota | 90.3 | 91.5 | 94.8 | 95.4 | 93 | 94.1 |
| Mississippi | 93.9 | 93.8 | 94.7 | 95.2 | 93.1 | 94.3 |
| Nevada | 76.0 | 76.9 | 77.3 | 79.0 | 72.7 | 75.1 |
| Oregon | 65.0 | 71.5 | 75.6 | 77.7 | 72.3 | 75.3 |
| Pennsylvania | 73.1 | 71.8 | 68.8 | 71.0 | 67.1 | 69.6 |
| Virginia | 86.0 | 88.0 | 82.7 | 85.0 | 77.7 | 83.0 |
| Washington | 48.3 | 55.5 | 65.7 | 67.5 | 63.6 | 66.7 |

**Table 4 Accuracy of Decision Tree, Bayesian Network and SVM models on Different States (3 attributes)**

| Attributes Used | Maximum span length, Average span length, Deck structure type | | | | | |
|---|---|---|---|---|---|---|
| **State** | **BN (%)** | | **DT (%)** | | **SVM (%)** | |
| | **Precision** | **Recall** | **Precision** | **Recall** | **Precision** | **Recall** |
| California | 67.4 | 68.2 | 68.0 | 70.5 | 68.3 | 69.8 |
| Florida | 83.7 | 85.3 | 76.3 | 76.9 | 82.2 | 86.0 |
| Georgia | 88.8 | 88.7 | 85.9 | 84.7 | 68.1 | 79.1 |
| Illinois | 82.3 | 82.9 | 64.1 | 76.9 | 72.4 | 80.8 |
| Maine | 67.5 | 71.8 | 72.7 | 77.0 | 69.9 | 76.4 |
| Minnesota | 84.6 | 85.0 | 88.0 | 89.5 | 86.0 | 88.7 |
| Mississippi | 87.4 | 87.9 | 90.7 | 91.7 | 88.4 | 90.6 |
| Nevada | 61.0 | 68.0 | 60.0 | 70.9 | 69.5 | 71.4 |
| Oregon | 69.9 | 67.4 | 71.3 | 73.8 | 72.3 | 75.7 |
| Pennsylvania | 53.2 | 52.5 | 54.4 | 55.6 | 34.4 | 49.6 |
| Virginia | 81.3 | 83.4 | 79.3 | 81.9 | 67.4 | 78.2 |
| Washington | 50.6 | 56.6 | 57.0 | 60.3 | 53.6 | 60.0 |

**Table 5 Reductions in prediction accuracy when material type is not considered**

| State | BN (%) | | DT (%) | | SVM (%) | |
|---|---|---|---|---|---|---|
| | Precision | Recall | Precision | Recall | Precision | Recall |
| California | 6.5 | 7.7 | 9.7 | 8.9 | 8 | 8.5 |
| Florida | -2.7 | 0.7 | 5.7 | 6.8 | -7 | -4.2 |
| Georgia | 7.4 | 8.2 | 9.6 | 11.5 | 22.4 | 14.2 |
| Illinois | 8.1 | 9.2 | 24.6 | 14.1 | 12.8 | 8.6 |
| Maine | 11.7 | 11.3 | 7.7 | 7.2 | 6.3 | 6.8 |
| Minnesota | 5.7 | 6.5 | 6.8 | 5.9 | 7 | 5.4 |
| Mississippi | 6.5 | 5.9 | 4 | 3.5 | 4.7 | 3.7 |
| Nevada | 15 | 8.9 | 17.3 | 8.1 | 3.2 | 3.7 |
| Oregon | -4.9 | 4.1 | 4.3 | 3.9 | 0 | -0.4 |
| Pennsylvania | 19.9 | 19.3 | 14.4 | 15.4 | 32.7 | 20 |
| Virginia | 4.7 | 4.6 | 3.4 | 3.1 | 10.3 | 4.8 |
| Washington | -2.3 | -1.1 | 8.7 | 7.2 | 10 | 6.7 |

**Table 6 Change in DT and BN model accuracies with seismic attribute compared to 4 attribute results**

| Attributes Used | Material type, Maximum span length, Average span length, Deck structure type, Seismic activity | | | | | |
|---|---|---|---|---|---|---|
| **State** | **BN (%)** | | **DT (%)** | | **SVM (%)** | |
| | **Precision** | **Recall** | **Precision** | **Recall** | **Precision** | **Recall** |
| California | 1 | -0.6 | 0.6 | 0.3 | 0.6 | 0.5 |
| Florida | 7.4 | 2.3 | 4.2 | 6 | 0 | -0.3 |
| Georgia | -0.3 | -0.5 | 0.4 | 0.4 | 0 | 0.1 |
| Illinois | 0.8 | 0.2 | 0.9 | 0.1 | 0 | 0 |
| Maine | -0.9 | -1.4 | 0.4 | 0.2 | 0.1 | 0.1 |
| Minnesota | 3 | 1.8 | 0.1 | 0 | -0.1 | 0.1 |
| Mississippi | 0.1 | 0.2 | 0.7 | 0.5 | 0.5 | 0.4 |
| Nevada | 1.8 | 0.2 | 0.8 | 1.1 | -0.9 | -0.6 |
| Oregon | 7.9 | -1.2 | 2.1 | 1.5 | 1.1 | 1.8 |
| Pennsylvania | -2.7 | -1.4 | 5 | 3.3 | 5.5 | 3.7 |
| Virginia | 1 | 0.3 | 4.2 | 3.5 | 0 | 0 |
| Washington | 16.4 | 8.9 | 2.3 | 2.5 | 0.4 | 1 |

**Table 7 Prediction accuracy for three common design types with and without seismic data (South Carolina, DT classifier)**

| Design Type | Without seismic attribute | Including seismic attribute |
|---|---|---|
| Slab | 87.8% | 92.7% |
| Stringer/Multi-beam or Girder | 73.1% | 94.6% |
| Culvert | 92.6% | 88.6% |

**Table 8 Change in DT, BN, and SVM model accuracies on resampled data compared to 4 attribute results**

| Attributes Used | Material type, Maximum span length, Average span length, Deck structure type, Seismic activity | | | | | |
|---|---|---|---|---|---|---|
| **State** | **BN (%)** | | **DT (%)** | | **SVM (%)** | |
| | **Precision** | **Recall** | **Precision** | **Recall** | **Precision** | **Recall** |
| California | 1.4 | -0.4 | 7.3 | 5.9 | 0.3 | -1.1 |
| Florida | 9 | 3.7 | 7.8 | 6 | -9.3 | -5.9 |
| Georgia | -0.8 | -1.3 | -1.6 | -1.2 | -9.2 | -5.8 |
| Illinois | 0.3 | -0.7 | 1.6 | -0.4 | -6.7 | -5.1 |
| Maine | -2.1 | -3.1 | 7.2 | 3.8 | 2.7 | -1 |
| Minnesota | 2.6 | 1 | 1.7 | 1.1 | -2.3 | -2.2 |
| Mississippi | -0.1 | -0.3 | 2.4 | 1.9 | 0.2 | -1.5 |
| Nevada | 3.7 | 2.1 | 10.8 | 8.9 | 2.2 | 1.8 |
| Oregon | 9.7 | 1.4 | 9.3 | 7.4 | 6.2 | 3.2 |
| Pennsylvania | -3 | -1.8 | 11.9 | 10.1 | 6.8 | 3.2 |
| Virginia | 1.4 | -0.3 | 4.9 | 3 | -6.7 | -4.2 |
| Washington | 15.3 | 8.6 | 14 | 12.6 | 3.7 | 11.5 |